\documentclass{article}
\usepackage{ijcai20}

\usepackage{times}
\usepackage{soul}
\usepackage{url}
\usepackage[hidelinks]{hyperref}
\usepackage[utf8]{inputenc}
\usepackage[small]{caption}
\usepackage{graphicx}
\usepackage{amsmath}
\usepackage{amsthm}
\usepackage{booktabs}
\usepackage{algorithm}
\usepackage{algorithmic}
\title{Cognitive Perspectives on Context-based Decisions and Explanations\thanks{The work is partially supported by the Wallenberg AI, Autonomous Systems and Software Program (WASP) funded by the Knut and Alice Wallenberg Foundation.}}

\author{
Marcus Westberg
\and
Kary Främling\
\affiliations
Computer Science Department, Umeå University, Sweden\\

\emails
marcus.westberg@umu.se,
kary.framling@umu.se
}

\begin{document}

\maketitle

\begin{abstract}
  When human cognition is modeled in Philosophy and Cognitive Science, there is a pervasive idea that humans employ mental representations in order to navigate the world and make predictions about outcomes of future actions. By understanding how these representational structures work, we not only understand more about human cognition but also gain a better understanding for how humans rationalise and explain decisions. This has an influencing effect on explainable AI, where the goal is to provide explanations of computer decision-making for a human audience. We show that the Contextual Importance and Utility method for XAI share an overlap with the current new wave of action-oriented predictive representational structures, in ways that makes CIU a reliable tool for creating explanations that humans can relate to and trust.
\end{abstract}

\section{Introduction}
Both making a decision and explaining a decision involves the structuring of concepts to provide a model for the context within which the decision took place and how the decision changed the factors present in this contest. For example, searching for a pair of scissors in a kitchen can involve the decision to open a drawer. The decision involves the application of the concept of 'scissors', 'drawer' as well as their relationship to each other. The localised understanding that scissors tend to be in kitchen drawers and that drawers are containers that can be pulled to have their contents revealed establishes affordances which aid the world navigation process. These applications of concepts need not be part of deliberate reasoning but could rather be an intuitive reaction as the result of ingrained patterns of knowledge and action, as suggested by \cite{Kahneman2011}. Explanations of decisions in turn attempt to retread these deliberations, or create post-hoc narratives to explain deliberations hidden to us. Similar approaches in relation to eXplainable AI (XAI) have been explored by \cite{miller2019}, based on the idea of explanation as conversation \cite{Hilton1990}, where the task is to "resolve a puzzle in the explainee's mind about why the event happened by closing a gap in his or her knowledge." In this paper we argue that explanations, while not always perfectly accurate in regards to reality (due to the existence of hidden factors), are best structured and involve the same conceptual basis as decisions, and that when trying to understand an explanation we do so by simulating the decision-making process through the explanation provided. In other words, what makes a good explanation of an agent-based action is that it presents us with a reasoning structure that we can follow and relate to our own decision-making processes. It is thus imperative that the explanations provided by artificial agents, in the context of XAI, not only provide deliberations that we can follow, but more importantly provide them in a conceptual framework which facilitates retreading the deliberation and is context-sensitive. 

When looking for a method to provide explanations of AI decisions, it is thus imperative that the method employed is one that produces explanations that are meaningful. Contextual Importance and Utility (CIU) as developed by \cite{Framling1995,Framling1996} provides a method for explaining decisions that utilises context-sensitivity. The benefit of CIU is that it provides a model-neutral approach to XAI and, as we will argue in this paper, complements current trends in cognitive philosophy which utilise both predictive models and action-based processes in the form of embodiment or action-oriented representations to explain cognitive processes. In HCI and XAI, parity between artificial processes and human processes can be beneficial for engendering trust and encouraging interaction, as well as opening up the possibility of collaborative or mutualistic developments \cite{Westberg2019}. In the following sections we will have a look at the evolution of (mental) representations in both cognition and computation. We will show how CIU can be linked to current theories of world navigation and decision-making. 

\subsection{Importance and Utility}
CIU employs the concepts of Importance and Ulility as a method of explaining AI decisions. Importance here refers to the importance of an input, while Utility refers to the value of an input in terms of how well it approximates the desired criteria. Both these concepts are employed on a contextual basis as Contextual Importance (CI) and Contextual Utility (CU), together forming CIU. Formally CI is defined as follows:

\begin{equation}
CI_{j}(\vec{C},\{i\})=\frac{Cmax_{j}(\vec{C},\{i\})-Cmin_{j}(\vec{C},\{i\})}{absmax_{j}-absmin_{j} 
} 
\end{equation}

where $\{i\}$ is a set of inputs and $j$ is a specific output in the context of $\vec{C}$. Consequently CU is formally defined as:

\begin{equation}
CU_{j}(\vec{C},\{i\})=\frac{out_{j}(\vec{C})-Cmin_{j}(\vec{C},\{i\})}{Cmax_{j}(\vec{C},\{i\})-Cmin_{j}(\vec{C},\{i\})} 
\end{equation}

When answering questions such as "why did you do \emph{X}" or "why did you not choose \emph{Y}", CIU evaluates decisions in explanation by looking at the importance and utility of factors in the decision-making process. Both importance and utility are context-sensitive and may change depending on the situation, but for any given context the definition of importance and utility are as follows:

Importance highlights what factors are important for a given decision. When deciding which pair of trousers to buy in a store, size will definitely be an important factor, as may design and price. After all, we want trousers that fit us\footnote{This is of course also context-sensitive. Oversized trousers would gain importance in the context of a clown buying trousers for a show.}, design that we like or feel comfortable in, and a price that we find affordable. The utility values of a given pair of trouser's size, design and price determine how well these individual factors fulfil the criteria of what we want out of our decision (so for a person with a 30 inch waist, trousers of waist size 30 would carry a high utility value). Utility in this sense is not linear, because trousers that are too large and too small both have low utility values, while a pair that fit perfectly have a very high utility value.

\section{Importance and Utility in Predictive Vision}
Predictive vision in philosophy \cite{clark2015,clark2016,hohwy2013} and cognitive science \cite{Hinton2007,Friston2008,Knill2005,Moreno-Bote2011} promotes a pro-active account of perception where mind and world meet halfway in order to form a more efficient way of gathering and sorting input signals, i.e. information about the world. This is done by creating an internal representational model of the world through which an individual forms predictions about future input given a certain planned or expected adjustment to the world (either through movement of the body through the world, thus adjusting perceptual perspective, or through modifying aspects of the world by moving objects or making other alterations). These predictions are then matched with the incoming input, comparing the expectations (seeing a pair of scissors then opening a drawer) to the result. Error signals are then generated where expectations and reality mismatch (absence of a pair of scissors would generate a much stronger error signal than the scissors being in an odd or otherwise unexpected position within the drawer). These error signals are then explained away by updating our understanding of the world and creating a new model (hypothesis) for where to find the scissors. In this way, most kinds of decision-making involve this type of prediction-based model generation. The concepts of importance and utility can be easily incorporated into predictive vision to highlight how these concepts operate.

When looking for a pair of scissors, I model my expectations of the scissors being in the kitchen, inside a drawer. To locate the scissors I will navigate into the kitchen and start opening drawers, looking for scissor-shapes. When doing so the importance and utility of the visual input changes. My visual processing will prioritize focus on scissor-like shapes (pointy object with two round holes) and objects that have features in common with my scissors. Thus the importance of such objects in my visual field will rise, their weights will be greater and my focus will be drawn to them. If my pair of scissors have an orange handle, then orange objects will also gain some amount of importance. Objects that fit all the criteria will then gain the highest weight of all, and thus become the anchoring of my focus when found.

When choosing between options during a decision process, utility values come into play. Again, these can be predictive in the sense that we have a (detailed or approximate) model for the ideal candidate in our mind. We may have an idea of what the ideal size of a mug is when we're looking for one, and depending on the type of beverage and how much we want of it, the ideal mug size will vary. For example, a cup of latte and a cup of espresso have very different ideal size values. In our decision-making process our focus will thus disregard mugs that fall too far outside the ideal range, i.e. are too big or too small. This range may alter depending on our available options. For example, if when looking through the cupboard we realise that all the big mugs are missing, we may instead start looking at the biggest available out of the small mugs. Alternatively we may adjust our world model and start looking for the ideal mug in the dish rack or dishwasher. Our decision capacity is thus not split into separate local importance values of a latte-context mug to be of size \emph{x} and an espresso-context cup to be of size \emph{y}, but rather we exploit a global importance of mug size in beverage contexts that is regulated by the contextual utility values, allowing for flexibility when reevaluating the environment.

\section{Representational Structure in Action}
A representative system is defined by \cite{Haugeland1991} as follows: In acting upon the environment, the system makes use of features that are not always reliably present to the system. What this means is that when navigating the world, the system makes use of features that it is aware of but that are not in the system's immediate vicinity (i.e. no input signals of such a feature are currently accessible to the system). For example, when a human decides to walk to a museum, the human's cognitive system will coordinate its navigation with the environmental feature of a museum (including location of the nearest museum, or a specific one depending on context) even if the museum is physically out of reach and view. In turn this navigation will employ other features such as known paths to the museum, turns to take etc. In order to do so, there must be something that the system can make use of that stands in for the actual input signal of a museum. This stand-in feature is a mental representation. By employing a representation of a museum, a human can think about and make decisions involving museums without actively perceiving a museum. By contrast, flowers do not need a representation of the sun in order to track it with their leaves and adjust for optimal sunlight intake, for whenever such coordination with the environment is happening the input signals (sun rays) are present. Haugeland further states that for a system to be representational, these stand-in processes need to be part of a systematic approach where representations are employed in a variety of states, such that a representation is not unique to a singular situation but can be employed in general ways, and that the representations carry meaning within the system in such a way that they could be employed improperly (mistakenly going to Anna's house when intending to go to Hanna's house) or misrepresent (thinking of the moon as being made of cheese).

While this constitutes a general account of what representations \emph{are}, the proposed structure and dynamics of representations has been iterated upon from their original conception, and old iterations are still relevant in certain contexts and applications. We will go through the most prominent accounts of representations, starting with what could be referred to as the 'classical' account and then follow the growing influence of embodiment in cognition.

\subsection{Classical representational systems}
Historically speaking, it is difficult to determine the exact origin of what can be classified as the classical form of representational systems. On a contemporary scale, a good example of classical representations can be found in \cite{fodor1975,Fodor1981representations}. Classical representational systems are primarily focused on globally effective representations possessing general properties that define a class of objects or concepts that require representing. In a classical system, searching for or thinking about a pair of scissors involves deploying a representational concept that structurally consists of the features an object would possess that identifies it as a pair of scissors. This would involve shape, feel, capacity etc. \cite{clark1997}. Thus, when thinking about a pair of scissors, we would be thinking about an object with two sharp blades, two ring-shaped holes, a capacity to be pulled apart and pushed together as a means of cutting, and so on. More abstract representations could involve entities such as "Wednesday" or even "next Wednesday". Representations are employed in thinking about the things that they relate to, and does so globally in that they apply equally in all contexts involving them. The worry about a deadline next Wednesday employs the representation of "Wednesday" just as much as the happy memories of the date last Wednesday, or the plan to go to the doctor on Wednesday the 23rd. 

There is an enduring debate over how the content of representations is determined. There are internalists who view the content of a mental representation to be private and dependent on individually intrinsic properties, and there are externalists \cite{Burge1979,Putnam1975mom} who view representational content as being public and determined (to varying extents) by environmental factors. To an externalist, one person who believes that the moon is made of cheese and another person who believes the moon to be made of rock would still possess the same representation of "moon", because this representation is informed by the combination of natural (the moon in the sky) and social (interactions with others talking about or otherwise referring to the moon) environmental factors. The fact that their beliefs are different is simply due to one of these individuals being misinformed (and thus misrepresenting). By contrast, an internalist would argue that the representational contents would be different in virtue of these different beliefs about the composition of the moon.

\subsection{Personalised animate vision}
While the classical account for representational systems provides a robust model for thinking about things in the world, its proposed global structure seems inefficient when it comes to picking out specific objects from a group. A person may be looking for their mug on a shelf full of other mugs. In this context, many objects in the visual field can be identified as mugs as per the representation. This person's mug may be blue, giving it a unique qualifier, but even then the mug's feature of being blue is a relatively minor feature of the global representation "my mug", which also includes all distinctive features identifying the representation "mug". Following this, it seems that the majority of computational effort spent identifying "mug" features is redundant. Research on animate vision systems by \cite{Ballard1991} provided an alternative to the theory on visual representations proposed by \cite{marr1980,Marr1982} and produced a more efficient approach to computational load in vision, involving personalised local representations. These personalised representations put extra emphasis on features that are helpful in local identification, but not necessarily in global identification. For example, in the representation of "my mug" the importance of the colour blue may be much greater for the identification process than other traits common to all mugs, even if the feature of colour itself has low importance in identifying an object as a mug. This puts personalised, or agent-dependent, representations in animate vision in stark contrast with agent-independent properties of classical systems. In this sense, personalised representations employ context-sensitivity: the representation of "my mug is blue" is only effective in the context of looking for one's mug, but it is also preferable over global descriptions during search tasks and, especially, in communication and explanation. After all, when we produce explanations of our world navigation process, for example answering the question "what are you looking for", we are unlikely to state "I'm looking for my mug" and then proceed to describe general mug descriptions, instead our explanatory focus is on the personalised aspects such as "I'm looking for my mug, it's blue". It is also important to note that even if personalised animate vision representations produce a stronger alternative to classical representations in some contexts, this does not make the latter obsolete. In fact, a complete account of human cognitive capacities involving representations would likely employ both global and local ways of representing, as purely localised methods fall short when it comes to context-neutral or open-ended tasks \cite{clark1997}.

\subsection{Action-oriented representations}
\cite{clark1997} argues that the complex internal states that constitute representations in some way function as specific information carriers through their correlation both within the system as well as with the body and world, suggesting that representations may be entangled with the world through action. Building upon ideas of embodied couplings with the world \cite{brooks91a,brooks91b,varela1991}, action-oriented representations stress the entanglement between mind, body and world by presenting representational structure that not only describes the world but prescribes possible actions and responses \cite{clark2016}. Instead of perception being data set up for exploitation by independent action systems, action-oriented representations create a representational bridge between perception and action. 

Clark argues that the Mataric robot \cite{mataric1990,mataric1992}, which registers landmarks in a maze as a combination of sensory input and current motion, displays the use of exactly these types of representations. A narrow corridor represents both the visual image of the corridor, as well as “forward motion”. As the robot creates a map of the maze, it is thus full of visual information as well as recipes for action \cite{clark2015}.

A football can be represented as a round object of certain hardness and visual pattern, but in action-oriented terms a football can also be represented as "affords kicking". Furthermore, contextual weights can determine what kind of actions are appropriate in a given environment. During a game a football, the prescribed action structure would involve "passing", "dribbling" and "scoring", while in the context of friendly play or a game without rules these action affordances may be much more general or basic. Additionally, certain contexts may change prescriptive actions to proscriptive, such as "pick up" or "throw" during a game with enforced no-hands rules. Such prescriptions and proscriptions, and the recognition of contextual shifts, are part of a learning process where the dynamic bonds between mind and environment are formed. Individuals who are unfamiliar or new to the context of a football game may during play momentarily regress to a more global set of affordances and perform a proscriptive action causing a foul (such as blocking a suddenly incoming ball with their hands) and then exclaim "oh, I forgot!". 

Just like personalised representations above, action-oriented representations also show descriptive power in communication and explanation. For example, when Dan is asked why he gave Stacy a glass of water, an appropriate answer would be "because she was thirsty" or "so that she could have a drink". This refers to the action affordance of water to drink it, and its capacity to slake thirst, thus the important features of water in this explanatory description in this context are action-oriented. Even though it would be accurate to say "because water is a liquid and Stacy requires water to slake thirst", this is an inefficient and awkward explanation of the action. In the context of XAI, a robot providing the former type of explanation would generate far more social connection and understanding from a human user than the latter.

\subsection{Changing utility in affordances}
To put these affordances in terms of contextual importance and utility, the importance of the "pick up" action may be very high during a game of football, with its utility being very low, i.e. it is very important not to pick up the ball. Meanwhile, while passing around a ball in a backyard with friends, the importance of "pick up" would decrease because of a lack of purpose or benefit of this action. Incidentally, the utility would also likely go up slightly because touching the ball with one's hands is neither a negative (proscribed) act nor especially beneficial. Through this perspective, CIU perfectly illustrates in these cases how the weights system of representational structures may alter between contexts, in a way that makes sense for human behaviour.

In american football, the affordances of throwability and catchability are very important when selecting a ball. However, "throwability" and "catchability" are intermediate concepts made up of more basic features that a ball can have, e.g. size, shape, inflatedness etc. \cite{Framling1996}. These basic values may change individually to affect the intermediate value. "Deflategate" was an NFL controversy involving a team using deflated balls in order to gain an advantage in play. Slightly deflated balls of 10.5 PSI are easier to grip compared to the rules-legal standard of 12.5 PSI. This means that the utility value for a 10.5 PSI ball is very good for inflatedness and thus also throwability, but in the context of wanting to follow the rules, this value is very bad. Thus when attempting to pick out a 'good' ball, the situation (context) has a great impact on how the intermediate concepts that make the ball 'good' are interpreted.

\section{Conclusions}

The primary strength of CIU as a method of explanation in AI is the integration of utility into the decision-making narrative. This gives it the upper hand over exclusively importance-focused methods such as LIME \cite{ribeiro2016i} because, as shown in this paper, to create a complete account of the human decision-making process, representational accounts on human cognition not only recognise importance, but utility as well. The problem to solve about AI is not only a computational problem, but a social one as well. Ultimately, AI interfaces with human lives, and as such it is important for communication between humans and AI to be meaningful. In terms of XAI, meaningful explanations carry a requirement of relatability to its audience. CIU is a proof-of-concept for the new cognitive landscape of context-sensitive representations in decision-making and world navigation, and fills in the mechanical gaps that philosophical theories don't fully provide. In turn, this means that CIU is not only a good method of explanation, but also a method of explanation that complements human mechanisms of understanding and organising the environment. From a philosophical standpoint, we argue that this makes CIU a strong candidate in XAI for communicating explanations not only to experts but for layperson sensibilities as well.

\bibliographystyle{named}
\bibliography{ijcai20}

\end{document}